\documentclass{article}

\usepackage{arxiv}

\usepackage[utf8]{inputenc} 
\usepackage[T1]{fontenc}    
\usepackage{hyperref}       
\usepackage{url}            
\usepackage{booktabs}       
\usepackage{amsfonts}       
\usepackage{nicefrac}       
\usepackage{microtype}      
\usepackage{lipsum}		
\usepackage{graphicx}
\usepackage[numbers]{natbib}
\usepackage{doi}

\usepackage{amsmath}

\title{A Step Towards Mixture of Grader: Statistical Analysis of Existing Automatic Evaluation Metrics}

\author{ \href{https://orcid.org/0009-0004-5472-2006}{\includegraphics[scale=0.06]{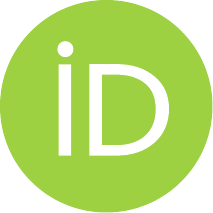}\hspace{1mm}Yun~Joon~Soh}\\
	Department of Computer Science\\
	University of California, San Diego\\
	San Diego, CA 92093 \\
	\texttt{yjsoh@ucsd.edu} \\
	\And
	\href{https://orcid.org/0000-0002-8766-0946}{\includegraphics[scale=0.06]{orcid.pdf}\hspace{1mm}Jishen~Zhao} \\
	Department of Computer Science\\
	University of California, San Diego\\
	San Diego, CA 92093 \\
	\texttt{jzhao@ucsd.edu} \\
}

\date{}

\renewcommand{\undertitle}{NeurIPS 2024 Workshop SFLLM}
\renewcommand{\shorttitle}{A Step Towards Mixture of Grader}

\begin{document}
\maketitle

\begin{abstract}
The explosion of open-sourced models and Question-Answering (QA) datasets emphasizes the importance of automated QA evaluation.
We studied the statistics of the existing evaluation metrics for a better understanding of their limitations.
By measuring the correlation coefficients of each evaluation metric concerning human-like evaluation score, we observed the following: (1) existing metrics have a high correlation among them concerning the question type (e.g., single word, single phrase, etc.), (2) no single metric can adequately estimate the human-like evaluation.
As a potential solution, we discuss how a Mixture Of Grader could potentially improve the auto QA evaluator quality.
\end{abstract}


\section{Introduction}

Large Language Models (LLMs) are widely adopted across various tasks including Question-Answering (QA) tasks.
More and more models including the fine-tuned models and the dataset used for fine-tuning are being released daily.
This explosion in the number of models, and datasets emphasizes the importance of accurate automatic evaluation for out-of-the-model language model training as well as gauging their QA capabilities.

However, varying question types (short-form, long-form, open-ended, etc.) and ambiguity in the grading rubric make it difficult to properly gauge each model's capability objectively for QA tasks.
No single existing evaluation metric can capture the language model's QA answer capability for multiple quality types.
For example, Exact Match (EM) is a widely adopted all-or-nothing evaluation metric that shows a high correlation with human-evaluated scores for short-form QA tasks but is too strict to give credit for any semantically identical answer.
The lack of an objective grading rubric for varying QA types creates a bias in summary statistics.
For example, half credit for an open-ended question is regarded equally as half credit for a simple factual question.


In this paper, we (1) deploy statistical approaches to characterize various existing evaluation metrics, (2) the effectiveness of recent ChatGPT-o1-preview model~\cite{chatgpto1} as QA grader, and (3) potential solution, a Mixture Of Grader (MOG), which first classifies each (question, gold answer) pair into one of the predefined QA type class and select the appropriate evaluation metric accordingly for an advanced automatic evaluation that better ``correlates'' to human evaluator.
\section{Background and Related Work}

Both industry and academia released various fine-tuned language models along with the dataset used for fine-tuning.
As some reports~\cite{Roy_2024} warn of the depletion of trainable data soon, generating quality datasets via LLMs is also gaining attention~\cite{li2024autobenchercreatingsalientnovel}.
With the explosion of the open-sourced models and datasets, the need for good quality QA evaluation metrics surfaces~\cite{adlakha2024evaluatingcorrectnessfaithfulnessinstructionfollowing}.

\subsection{Problem Statement}

We define the problem as follows; Given a list of ``gold'' answers (typically aggregated list of multiple human annotations) and an attempted answer, a \textit{QA evaluation metric} returns a real number, \texttt{score}, where $0 \leq score \leq 1$ and a score close to 0 would typically indicate that the attempted answer is likely to be incorrect.

\subsection{Automatic Evaluation Metrics}
Several automatic evaluation metrics have been proposed ranging from token-counting to agentic evaluation approach. 
We briefly describe them and their advantages and disadvantages.

\paragraph{Token Counting}
\textbf{Exact Match (EM)} is a discrete metric (either 0 or 1) that is often too strict for semantically correct answers. As a complementing metric, many works report \textbf{token-level F1 Score}. F1 score, a harmonic mean of precision and recall, is more permissive than EM giving a partial score ($score \in [0,1]$). On the other hand, F1 scores are misleading when a model includes an additional explanation as part of the attempted answer because the additional tokens reduce the precision and thus the F1 score. For this reason, several papers included just the \textbf{token-level recall} as an evaluation metric. The downside of the recall metric is that naively repeating the context for contextual QA would yield a high score as it does not get penalized for lengthy answers.

\paragraph{N-gram Metrics}
\textbf{Bilingual Evaluation Understudy (BLEU)} measures the n-gram overlap between the predicted answer and the gold answer.
It is calculated as the geometric mean of n-gram precision (i.e., true positive count / (true positive + false positive count)).
Unlike naive token counting, BLEU rewards a sequence of token overlaps capturing the partial correctness more objectively than token-level metrics.
\textbf{Recall-Oriented Understudy for Gisting Evaluation (ROUGE)} also leverages n-grams and measures how many n-grams in the correct answer are also present in the generated answer.
Many variants have been proposed and we use the most widely adopted version, which takes the Longest Common String instead of a naive n-gram.
The benefit of ROUGE-L is that it captures the sentence-level structural correctness, which may be difficult for BLEU.

\paragraph{Machine Learned Evaluation Metric}
Recent works leverage the machine-learned method as a QA grader~\cite{li2024pedantspreciseevaluationsdiverse,llama_index,dubois2024lengthcontrolledalpacaevalsimpleway,lin2023llmevalunifiedmultidimensionalautomatic}.
Despite their ability to better understand the evaluation instructions and return a score with logical justification, the evaluation quality fluctuates from model to model and prompt to prompt.
Furthermore, agentic evaluators are computationally expensive which may impact the overall training performance if the QA task evaluation is included as part of the training loop.
\section{Proposal}

Instead of searching for a single metric that can handle all QA types properly, we propose a Mixture Of Grader (MOG), which first classifies the question and answer pair and selects the appropriate evaluation metric based on the classification outcome.
As the first step, we study the strengths and weaknesses of existing metrics.
With the help of the ChagGPT-o1-preview model, we devised the QA types including a single word, numerical, paragraph, code snippet, sentence, equation, phrase, name, boolean, list, symbol, single character, formula, long paragraph, essay, and short paragraph.

\section{Preliminary Results}

We discuss the experiment setups, visualize the statistics result, and share our interpretation of the result.

\subsection{Experiment Setup}

We assume that the ChatGPT-o1-preview model performs human-like evaluation on a (question, gold answer, attempted answer) tuple and study the correlation between the evaluation metric and the ChatGPT's evaluation score.

\paragraph{Dataset}

We synthesized the dataset by asking the ChatGPT-o1-preview model to generate data samples of (question, gold answer, attempted answer, score, justification, question type, answer type) tuple.
A total of 359 data were generated with an average score of 0.42 and a standard deviation of 0.42.
Each data has 6 fields and we count the number of unique entries per field in Table~\ref{tab:ss_dataset}.

\begin{table}
    \centering
    \begin{tabular}{c|c|c|c|c|c}
        Question & Gold Answer & Attempt Answer & Justification & Question Type & Anser Type  \\
        \hline
        352 & 313 & 314 & 348 & 61 & 16
    \end{tabular}
    \caption{Unique element count in the synthesized dataset generated by ChatGPT-o1-preview}
    \label{tab:ss_dataset}
\end{table}

\paragraph{Evaluation Metrics}

\begin{figure}
\includegraphics[width=\linewidth]{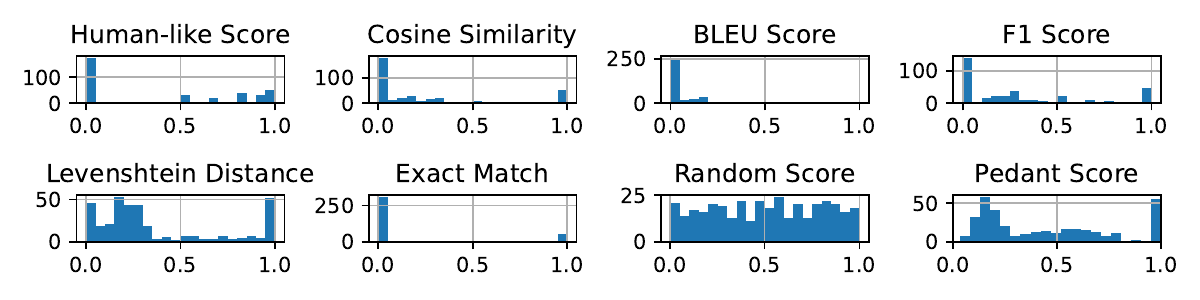}
\caption{Correlation Matrix Between Evaluation Metrics}
\label{fig:score_distribution}
\centering
\end{figure}

For each (gold answer, attempted answer) pair, we evaluate the quality using the following evaluation metrics: Exact Match (EM), F1 Score, Recall, BLEU, ROGUE-L, Levenstein Distance, Cosine Similarity (Tokenizer: TFidf), PEDANTS~\cite{li2024pedantspreciseevaluationsdiverse}.
We show the score distribution for each metric in Figure~\ref{fig:score_distribution}.

\subsection{Main Result - Pedant is the winner}

To quantify, how well each evaluation metric ``estimates'' the human-like evaluation, we measure the arithmetic mean of the absolute distance.

\[
\text{Mean of Score Delta} = \frac{\sum_{i=1}^{n}{abs(s_i - \hat{s_i})}}{n}
\]

\begin{figure}
\includegraphics[width=\linewidth]{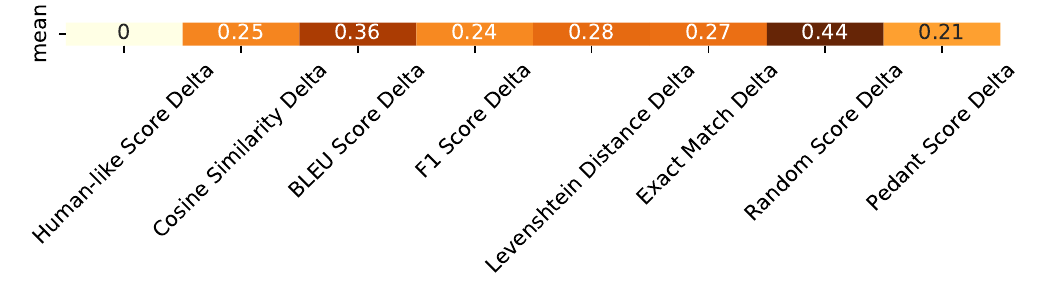}
\caption{Mean of Delta Scores i.e., abs(Metric Score - Human-lik Score)}
\label{fig:delta}
\centering
\end{figure}

\subsection{Comparison of Evaluation Metric}
\begin{figure}
\includegraphics[width=\linewidth]{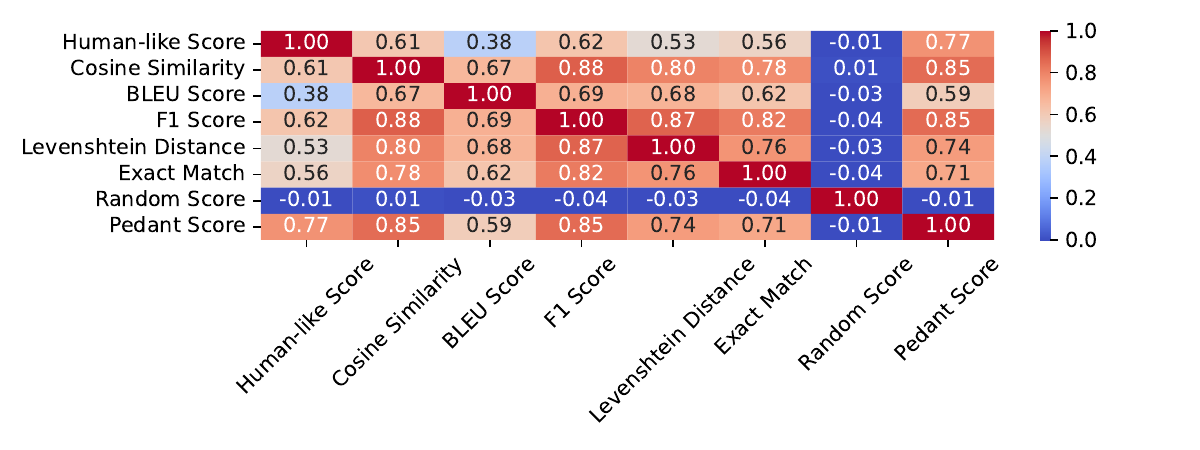}
\caption{Correlation Matrix Between Evaluation Metrics}
\label{fig:correlation}
\centering
\end{figure}

To better understand the difference between each evaluation metric, we measured the correlation of each evaluation metric score to the human-like evaluation score (Figure~\ref{fig:correlation}).
The key observation is that the Pedant score showed the highest correlation (0.77) with the human-like score.
This result aligns with the main result in (Figure~\ref{fig:delta}), where Pedant showed the least deviation from the human-like evaluation.
The observation also implies that a trained solution may work better than existing evaluation metrics and better estimate human-like evaluations.

From the correlation matrix, we further observed that many of these metrics show high correlations.
For example, the F1 Score showed a high resemblance to most evaluation metrics, supporting its popularity in various works.
As expected, the random scores were uncorrelated with any of the existing metrics.

\subsection{Per Answer Type Analysis}

\begin{figure}
\includegraphics[width=\linewidth]{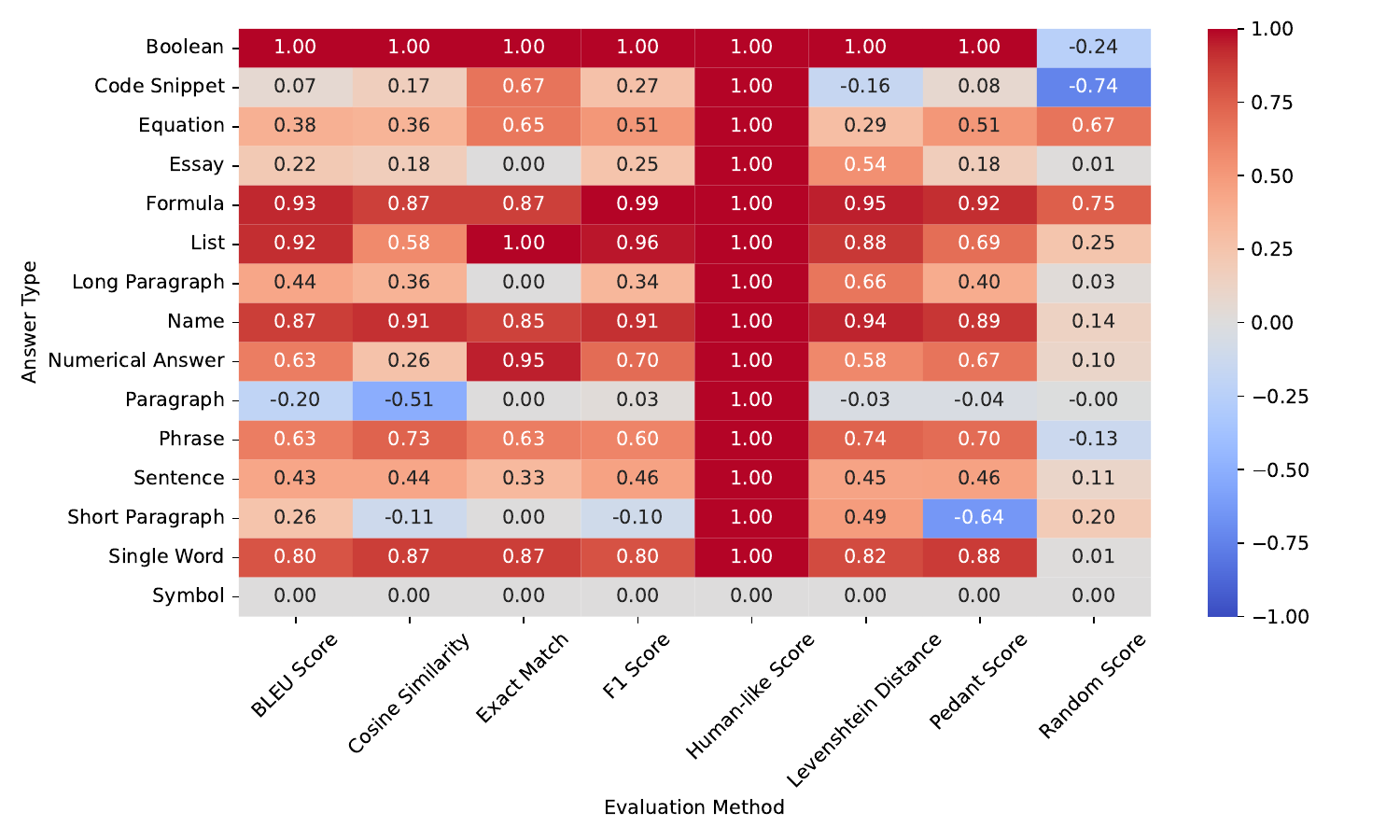}
\caption{Pearson Correlation Coefficient Matrix Per Answer Type Across Evaluation Metrics}
\label{fig:pearson}
\centering
\end{figure}

For deeper understanding, we measured per answer type (e.g., single word, paragraph, etc.) correlation coefficients.
Figure~\ref{fig:pearson} shows the Pearson Correlation Coefficient per answer type across multiple evaluation methods.
We make the following observations.

\begin{itemize}
    \item Exact Match (EM) was particularly strong for short-form answers such as single word, numerical, name, list, and formula, but not for long-forms such as essay, long paragraph, paragraph, short paragraph, or sentences.
    \item F1 Score showed a slightly lower coefficient than EM, but higher for answer types where EM suffered to correctly evaluate.
    \item Pedant performed worse than EM for short-forms but had a better correlation for long-form answers (essay or long paragraph) resulting in a better correlation in general (Figure~\ref{fig:correlation}).
\end{itemize}

From the observation, we conclude that there is no ``one-size fits all'' evaluation metric.
A better direction for a more concrete automatic evaluator would be to first classify the question and answer types and use different evaluation metrics for each type.
We left the details of question and answer type classification and per-type evaluation metric selection for future work.

\section{Discussion and Limitation}

We articulate the limitations of the current study.

\begin{itemize}
    \item The generated dataset has a limited number of samples for different answer types. This was mainly due to the limited query count to ChatGPT-o1-preview.
    \item The number of samples for each answer type varies significantly.
We left the proportion to be non-uniform as it may reflect the real-world scenario (i.e., users may ask a question that can be answered with a single word much more often than asking a question with a symbol as an answer).
    \item The TF-IDF tokenizer can be useful for many grading metrics but cannot properly vectorize symbols or equations. Trying with an alternative embedding model is left for future work.
    \item Several additional automatic evaluators~\cite{llama_index,dubois2024lengthcontrolledalpacaevalsimpleway} were recently proposed but left for future work.
\end{itemize}

\section{Conclusion and Future Work}

We explored the pros and cons of existing QA evaluation metrics concerning the score correlation to state-of-the-art LLM evaluator.
We show that each metric has strengths and weaknesses based on QA type and motivate a QA type classifier and QA type-based metric selection.

\newpage
\appendix

\section{ChatGPT-o1-preview's Justification for Scores}

Table~\ref{tab:dataset_just} shows some samples of ChatGPT's justification for giving the score.
It resembles how an actual human evaluator would give a partial score.

\begin{table}
    \centering
    \begin{tabular}{p{0.10\linewidth} | p{0.3\linewidth} | p{0.1\linewidth} | p{0.1\linewidth} | p{0.3\linewidth} }
        Question & Gold Answer & Attempted Answer & Score & Justification \\
        \hline
        "What is the capital of France?" & "Paris" & "Pariss" & 0.8 & "Minor spelling mistake but understandable." \\
        "Explain the process of photosynthesis." & "Photosynthesis is the process by which green plants convert sunlight into chemical energy." & "Plants make food using sunlight." & 0.7 & "Partial explanation lacking details about chemical energy conversion." \\
        "Is water a compound or an element?" & "Compound" & "Element" & 0.0 & "Incorrect; water is a compound composed of hydrogen and oxygen." \\
        "Who wrote 'Romeo and Juliet'?" & "William Shakespeare" & "William Shakespear" & 0.9 & "Minor spelling error but clearly identifies the correct author." \\
        "What is 15\% of 200?" & "30" & "20" & 0.0 & "Incorrect calculation; 15\% of 200 is 30." \\
    \end{tabular}
    \caption{Some Samples of ChatGPT-o1-preview Generated Dataset}
    \label{tab:dataset_just}
\end{table}

\section{Score Distribution Details}

\begin{figure}
\includegraphics[width=\linewidth]{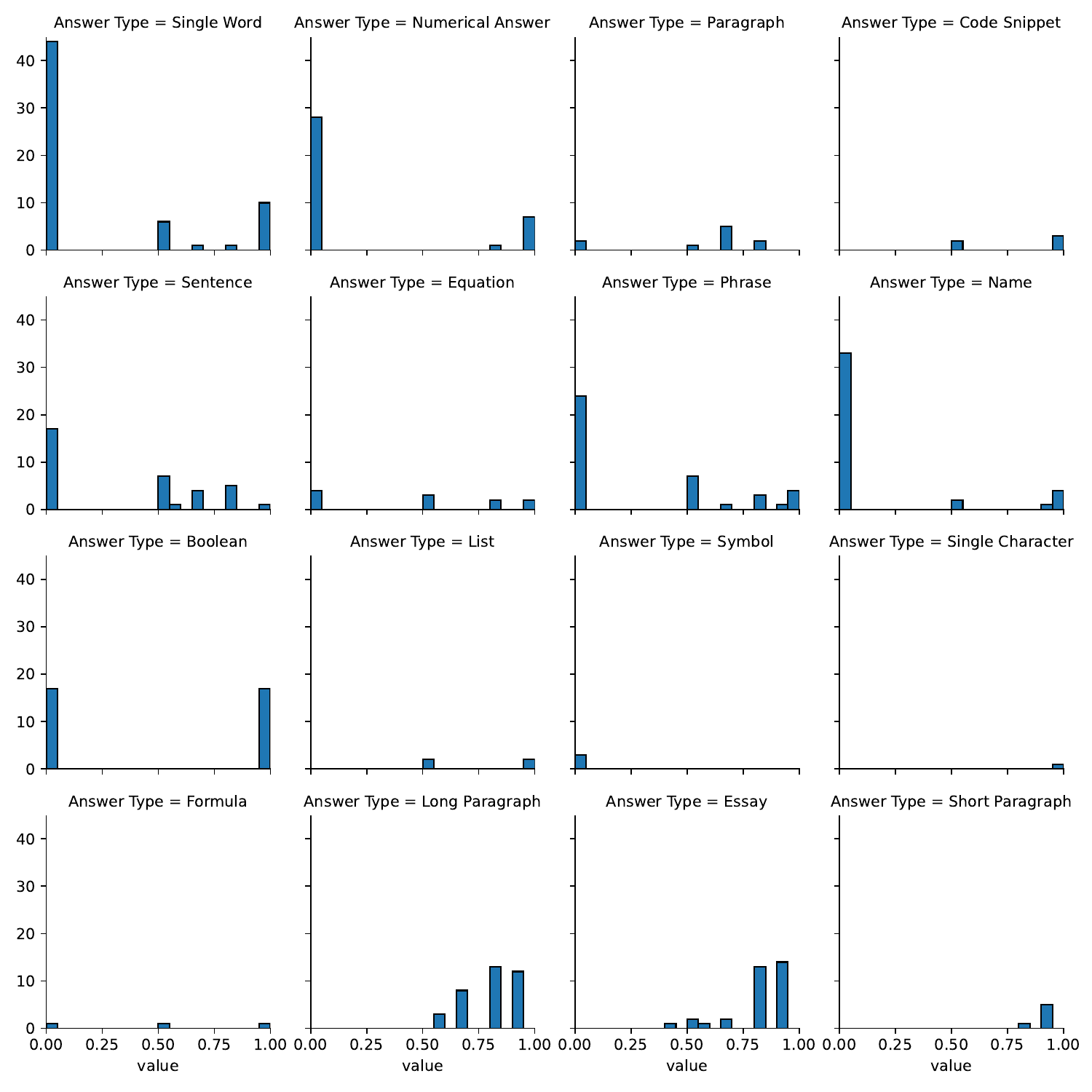}
\caption{Human-like Score Distribution Per Answer Type}
\label{fig:score_dist_all}
\centering
\end{figure}

Figure~\ref{fig:score_dist_all} shows the score score distribution per question type.

\section{Per Answer Type Mean Delta Score}

\begin{figure}
\includegraphics[width=\linewidth]{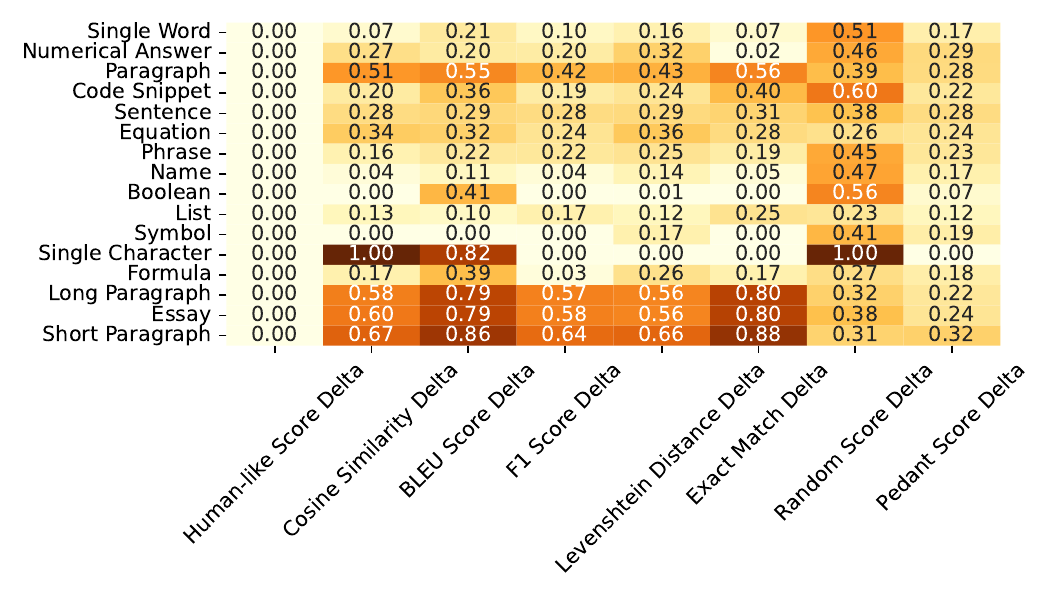}
\caption{Per Answer Type Mean Delta Score}
\label{fig:mean_delta_per_type}
\centering
\end{figure}

Figure~\ref{fig:mean_delta_per_type} is a per-answer type detailed version of Figure~\ref{fig:delta}.

\section{Additional Correlation Coefficient}

\begin{figure}
\includegraphics[width=\linewidth]{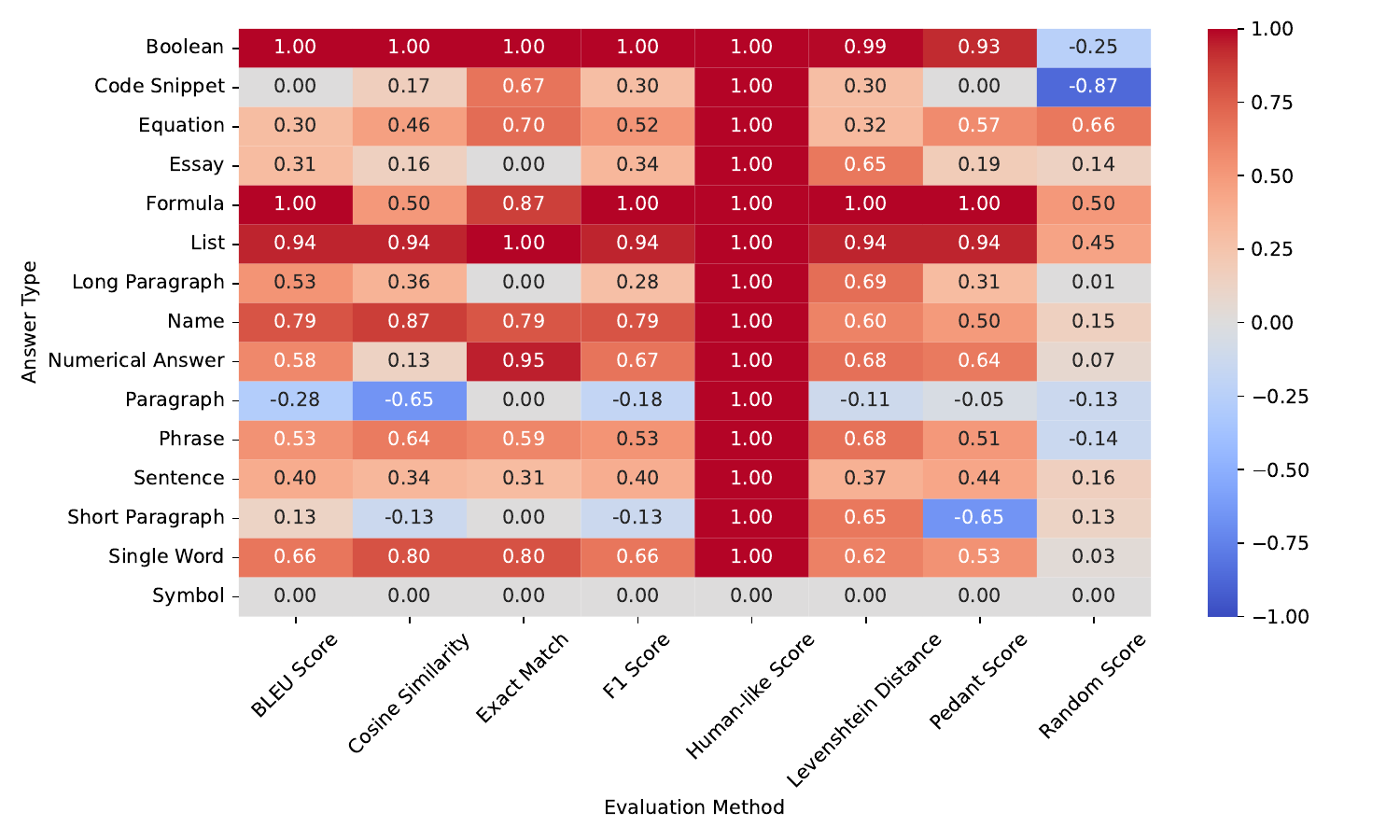}
\caption{Spearman Between Evaluation Metrics}
\label{fig:spearman}
\centering
\end{figure}

\begin{figure}
\includegraphics[width=\linewidth]{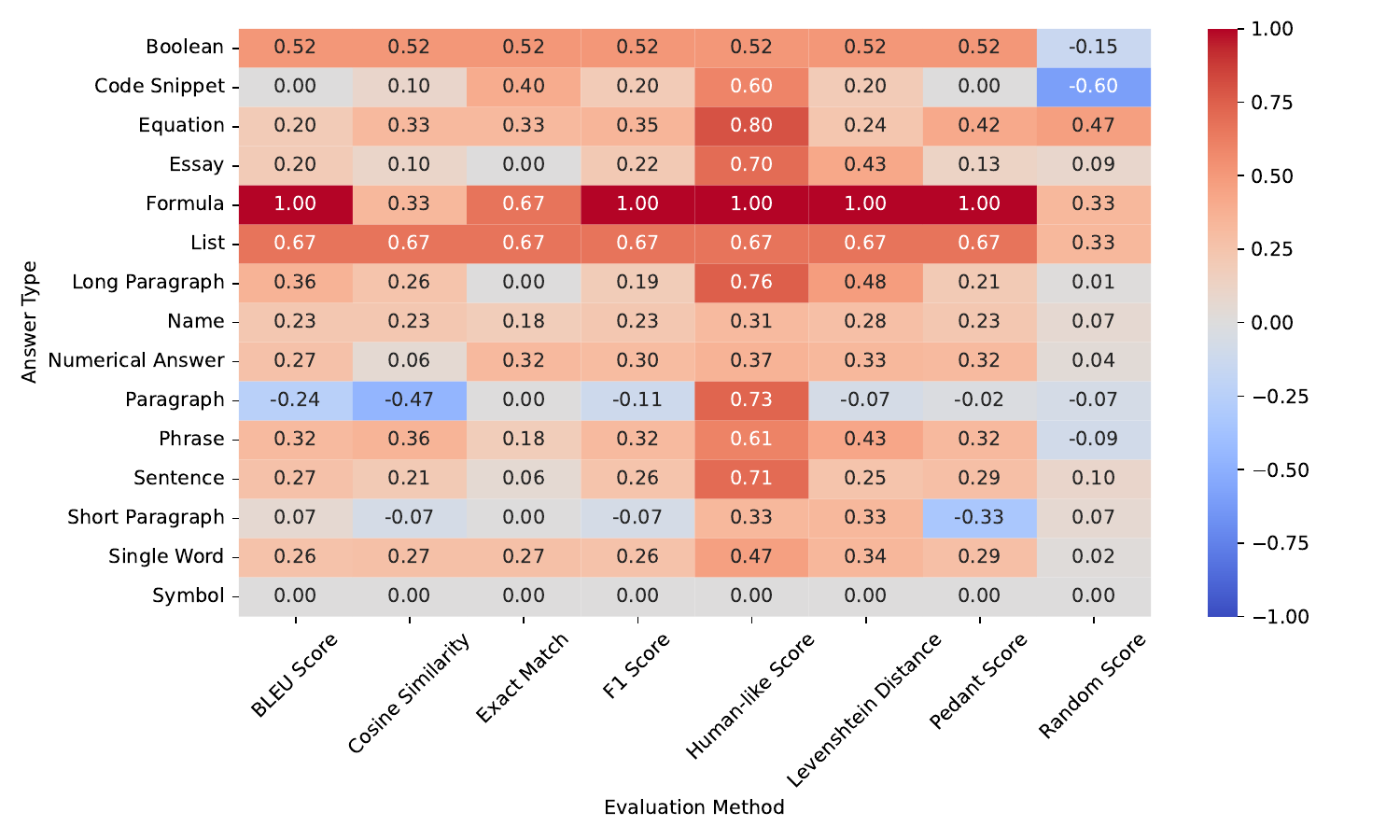}
\caption{Kendall $\tau$ Between Evaluation Metrics}
\label{fig:kendall}
\centering
\end{figure}

Although some prior works~\cite{li2024autobenchercreatingsalientnovel,adlakha2024evaluatingcorrectnessfaithfulnessinstructionfollowing} used the Spearman and/or Kendall's correlation coefficient, we moved these results (Figure~\ref{fig:spearman} and Figure~\ref{fig:kendall}, respectively) to appendix because these ranking based statistics were more suitable for ELO ranking based score as detailed in the prior work~\cite{li2024autobenchercreatingsalientnovel}.

\newpage
\bibliographystyle{plain}
\bibliography{paper}
\end{document}